\documentclass[runningheads]{llncs}
\usepackage{graphicx}
\usepackage[width=122mm,left=12mm,paperwidth=146mm,height=193mm,top=12mm,paperheight=217mm]{geometry}
\usepackage{booktabs}
\usepackage{multirow}
\newcommand{\etal}{\emph{et al.}}
\newcommand{\eg}{\emph{e.g. }}
\newcommand{\ie}{\emph{i.e. }}
\graphicspath{{./Imgs/}}

\begin{document}
\def\ECCV16SubNumber{}  

\title{Attentive Contexts for Object Detection} 



\author{Jianan~Li$^{* 1}$,
	Yunchao~Wei$^{* 2}$,
	Xiaodan~Liang$^{3}$,
	Jian~Dong$^{4}$,
	Tingfa~Xu$^{1}$,
	Jiashi~Feng$^{4}$,
	and Shuicheng~Yan$^{4}$
}

\institute{Beijing Institute of Technology University, China.
	\and
	Beijing Jiaotong University, China.
	\and
    Sun Yat-Sen University, China.
	\and
	National University of Singapore.
}

\maketitle

\vspace{-10mm}
\begin{abstract}
	Modern deep neural network based object detection methods typically classify  candidate proposals using  their interior features. However,  global  and  local surrounding contexts that are believed to be valuable for object detection are  not fully exploited by existing methods yet. 
	In this work, we take a step towards understanding what is a robust practice to extract and utilize contextual information to  facilitate object detection in practice. 
	Specifically, we consider the following two questions: ``how to identify  useful global contextual information for detecting a certain object?" and ``how to exploit local context surrounding a proposal for better inferring its contents?".
	We provide preliminary answers to these questions through developing a novel Attention to Context Convolution Neural Network (AC-CNN) based object detection model. AC-CNN effectively incorporates  global and local contextual information into the region-based CNN (\eg Fast RCNN) detection model and provides better object detection performance. It consists of one attention-based global contextualized (AGC) sub-network and one multi-scale local  contextualized (MLC) sub-network. To capture global context, the AGC sub-network recurrently generates an attention map for an input image to highlight useful global contextual locations,  through multiple stacked Long Short-Term Memory (LSTM) layers. For capturing  surrounding local context, the MLC sub-network exploits both the inside and outside contextual information of each specific proposal at multiple scales. The global and local context are then fused together for making the final decision for detection. Extensive experiments on PASCAL VOC 2007 and VOC 2012 well demonstrate the superiority of the proposed AC-CNN over well-established baselines. In particular, AC-CNN outperforms the popular Fast-RCNN by $2.0\%$  and $2.2\%$ on VOC 2007 and VOC 2012 in terms of mAP, respectively.

\end{abstract}

\vspace{-12mm}
\section{Introduction}
\vspace{-4mm}
Object detection  aims at localizing instances of real-world objects within an image automatically. The past few years have witnessed significant progress in this field, arguably benefiting from the rapid  development and wide application of deep convolutional neural network (CNN) models~\cite{krizhevsky2012imagenet,min2014network,szegedy2014going,simonyan2014very,he2015deep}. 

Among the CNN based object detectors, the Region-based Convolutional Neural Network (R-CNN) one~\cite{girshick2013rich} is seen as a milestone  and  has achieved state-of-the-art performance. The R-CNN model detects objects through using a deep CNN to classify region proposals that possibly contain objects~\cite{uijlings2013selective,zitnick2014edge,arbelaez2014multiscale}. Inspired by R-CNN, two follow-up models, \ie Fast R-CNN~\cite{girshick2015fast} and Faster R-CNN~\cite{ren2015faster}, have been developed to further improve the  detection accuracy as well as computational efficiency. Those two methods share a similar pipeline which becomes \emph{de facto} standard one nowadays: first apply Region-of-Interest (RoI) pooling  to extract proposal features from feature maps produced by CNN and then minimize  multi-task loss functions for simultaneous localization and classification. Many recent works~\cite{liang2015towards,liang2015reversible,zeng2015window,gidaris2015object} have also demonstrated the effectiveness of this pipeline. However, most of those state-of-the-art methods  localize objects using only information within a specific proposal that may be insufficient for accurately detecting challenging objects, such as the ones with low resolution, small scale or heavy occlusion.

It has been believed  for quite a long time that contextual information is beneficial for various visual recognition tasks. Many previous studies~\cite{choi2010exploiting,galleguillos2010context,oliva2007role,zweig2007exploiting,chen2015contextualizing,bell2015inside} have demonstrated considerable improvements for object detection brought by exploiting contextual information. For example, Chen \etal~\cite{chen2015contextualizing} proposed a contextualized SVM model for complementary object detection and classification, and they provided state-of-the-art results using hand-crafted features. More recently, Bell \etal~\cite{bell2015inside} presented a seminal work devoted to integrating contextual information into the Fast-RCNN framework. In~\cite{bell2015inside}, a spatial recurrent neural network was employed to model the  contextual information of interest around proposals. However, despite the fact that they provide new state-of-the-art performance, few of those methods present significant benefit of contextual information for object detection. How to effectively model and integrate  contextual information into current state-of-the-art detection pipelines is still  a worthwhile problem yet has not been fully studied. 

Intuitively, a global view on  background  of an image can provide useful contextual information. For example, if one wants to detect a specific \emph{car} within  a scene image, those objects such as \emph{person}, \emph{road} or another \emph{car} that usually co-occur with the target   may provide useful clues for detecting the target. 
However, not all the background information is positive for improving object detection performance --- incorporating meaningless background noise may even hurt the detection performance. Therefore, identifying useful contextual information is necessarily the first step towards effectively utilizing  context for object detection. In addition to such ``global'' context,  a local view into the neighborhood of one  object proposal region  can also provide some useful cues for inferring contents of a specific proposal. For example,  surrounding environment (\eg ``road'') and discriminative part of the  object (\eg  ``wheels'') could benefit detecting the object (\eg ``car'') obviously.

\begin{figure}[t]
	\centering
	\includegraphics[scale=0.43]{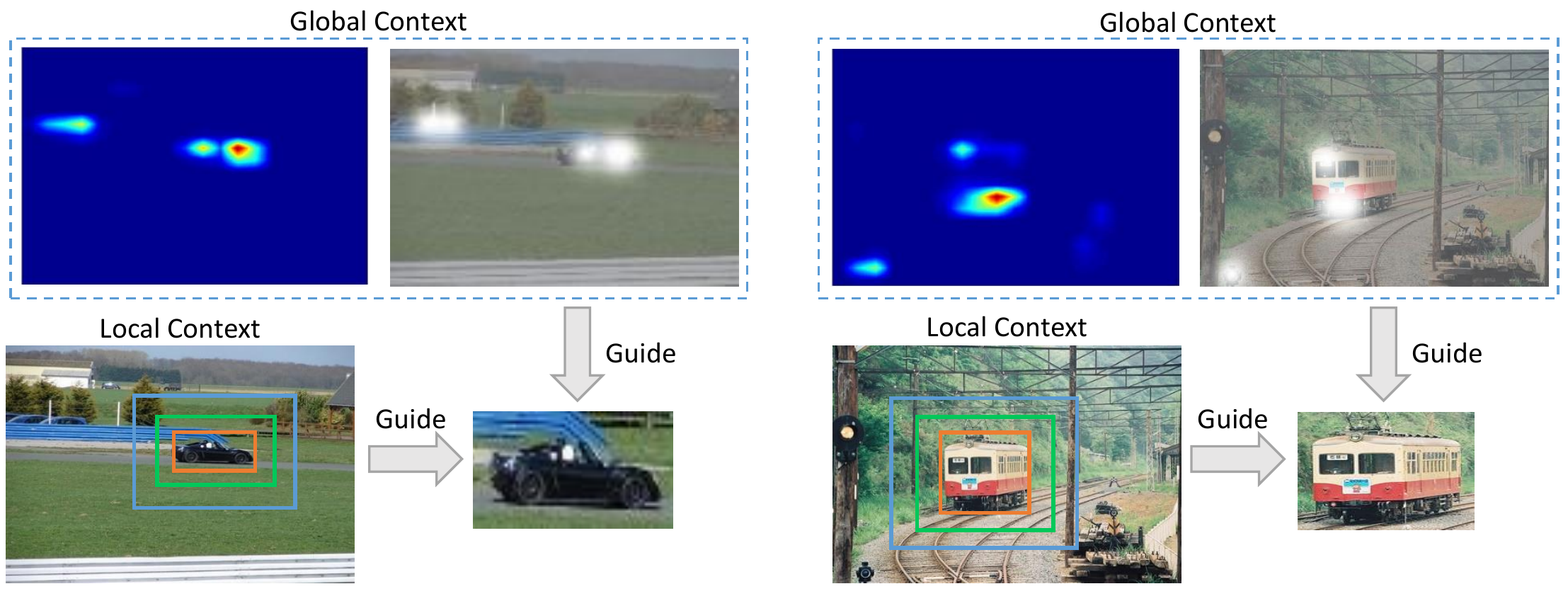}
	\caption{Illustration of incorporating both local and global contextual information for guiding object detection. For the local context, beyond the original bounding box of a specific proposal, inside and outside contextual information are employed to enhance the feature representation. For the global context, an attention based recurrent model is utilized to obtain positive contextual information (the highlighted region) from the global view.}
	\label{fig:motivation}
	\vspace{-7mm}
\end{figure} 

Motivated by the above observations, in this work, we propose a novel Attention to Context Convolution Neural Network (AC-CNN) model for better ``contextualizing'' the  region-based CNN detectors (such as Fast RCNN). AC-CNN captures contextual information from both global and local views, and effectively identifies helpful global context through  an attention mechanism. Taking the image  shown in Figure~\ref{fig:motivation} as an example, AC-CNN exploits local context inside  and surrounding a specific proposal  at multiple scales. As aforementioned, looking into the interior local context information helps discover some discriminative object parts and exploiting  external local context gives local cues  to determine what objects are present in the scene. Taking surrounding local context into consideration provides an enhanced feature representation of a specific proposal  for object recognition.


In addition, to identify useful contextual information from a global view, AC-CNN employs an attention-based  recurrent neural network that consists of multiple stacked Long Short-Memory (LSTM) layers to recurrently find locations within the input image valuable for object detection. As shown in Figure~\ref{fig:motivation}, the locations (around the target \emph{car} and another \emph{car}) with stronger semantic correlations  are discovered with higher confidences. In contrast, background noise that may hurt the detection performance is successfully suppressed on the attention map. Then, combining the feature maps of all locations guided by the attentive location map can produce  ``cleaned'' global contextual feature to assist  recognition of each proposal.


The main contributions of the proposed AC-CNN method can be summarized as follows:
\vspace{-4mm}
\begin{itemize}
	\item We propose a novel Attention to Context CNN (AC-CNN) object detection model which effectively contextualizes   popular region-based CNN detection models. To the best of our knowledge, this work is the first research attempt to detect objects by exploiting both local and global  context with attention.
	
	\item The attention-based contextualized sub-network in AC-CNN recurrently generates the attentive location map for each image that helps to incorporate the most discriminative global context into  local object detection.
	
	\item The inside and outside local contextual information for each proposal are captured by the proposed multi-scale contextualized sub-network.
	
	\item  Extensive experiments on the PASCAL VOC 2007 and VOC 2012 well demonstrate the effectiveness of the proposed AC-CNN: it outperforms the popular Fast-RCNN by $2.0\%$  and $2.2\%$ in mAP on VOC 2007 and VOC 2012  respectively.	
	We also visualize  results on automatically produced attention maps and seek to provide in-depth understanding on the role of global and local context for successful object detectors.
\end{itemize}	

\vspace{-4mm}
\section{Related Work}
\vspace{-2mm}
\subsection{CNN-based Object Detection Methods}
Object detection aims to recognize and localize each object instance with an accurate bounding box. Currently, most of the state-of-the-art detection pipelines follow the Region-based Convolutional Neural Network (R-CNN)~\cite{girshick2013rich}. In R-CNN, object proposals are first generated with some hand-crafted methods (\eg Selective Search~\cite{uijlings2013selective}, Edge Boxes~\cite{zitnick2014edge} and MCG~\cite{arbelaez2014multiscale}) from the input image, and then the classification and bounding box regression are performed to identity the target objects. However,  training an  R-CNN model is usually expensive in both space and time costs. Meanwhile, object classification and bounding box regression are implemented through a multi-stage pipeline. To enhance the computational efficiency and  detection accuracy, a Fast R-CNN framework~\cite{girshick2015fast} was proposed to jointly classify object proposals and refine their locations through multi-task learning. To further reduce the time of proposal generation, a novel Region Proposal Network (RPN)~\cite{ren2015faster} was proposed, which can be seamlessly embedded in the Fast R-CNN framework for proposal generation. However, those methods only consider  information extracted from a specific proposal in training, and  cues from context are not well exploited. In this work, we dedicate our efforts to attending contextual information based on the state-of-the-art pipeline to improve the accuracy of object detection. In particular, both local and global contextual information w.r.t.\ a specific proposal are taken into account for better object detection.

\vspace{-4mm}
\subsection{RNN-based Object Detection Methods}
\vspace{-2mm}
Recently, LSTM has shown  outstanding performance for the tasks of image captioning~\cite{xu2015show}, video description~\cite{yao2015describing,venugopalan2014translating}, people detection~\cite{stewart2015end} and action recognition~\cite{sharma2015action}, benefiting from its excellent ability to model long-range information. Most of those existing works  tend to adopt a CNN accompanied with several LSTMs to address specific visual recognition problems. Specifically, Sharma \etal~\cite{sharma2015action} proposed a soft visual attention model based on LSTMs for action recognition. Yao \etal~\cite{yao2015describing} proposed to use CNN features and an LSTM decoder to generate video descriptions. Stewart \etal~\cite{stewart2015end} employed a recurrent LSTM layer for  people detection. In this work, we offer the first research attempt to apply LSTM to learning the useful global contextual information with guidance from annotated class labels. Feature cubes of the entire image are taken as the input to a recurrent model consisting of multiple LSTM layers. With the recurrent model, some contextual slices beneficial for the detection task are iteratively highlighted to provide powerful feature representations for object detection.
\begin{figure}[t]
	\centering
	\includegraphics[scale=0.5]{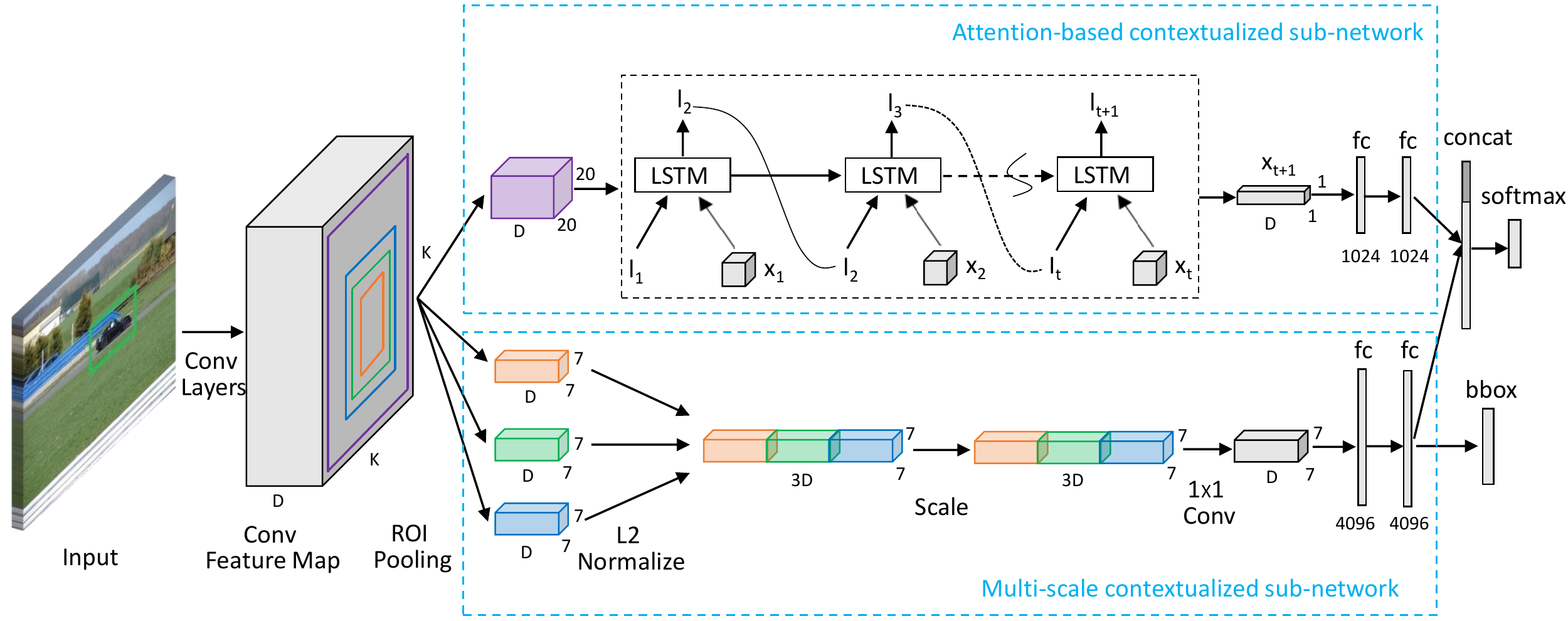}
	\caption{Details on how the proposed AC-CNN exploits contextual information for object detection. AC-CNN consists of two main sub-networks, \ie the attention-based contextualized sub-network and the multi-scale contextualized sub-network. An image is first fed into a convolutional network  to produce the feature cube. Then the feature cub passes through the multi-scale contextualized sub-network for local context information extraction. The bounding box of each proposal is scaled with three pre-defined factors and feature representations from the bounding boxes are extracted by an RoI pooling layer. Each feature representation, after L2-normalization, concatenation, scaling, dimension-reduction, is then fed into two fully-connected layers. In the attention-based contextualized sub-network, the feature cube is first pooled into a cube with a fixed scale. Then, a recurrent attention model including three LSTM layers  is adopted to recurrently detect useful regions from a global view. Finally, a global context feature is pooled based on the calculated attention map and fed into two fully-connected layers. AC-CNN uses  the output feature from the multi-scale contextualized sub-network for bounding box regression. The concatenated feature of the outputs from both two sub-networks is used for object classification.}
	\label{fig:framework}
    \vspace{-6mm}
\end{figure} 

\vspace{-2mm}
\section{Attention to Context Convolution Neural Network (AC-CNN)}
As shown in Figure~\ref{fig:framework}, the proposed AC-CNN method is built on the VGG-16 ImageNet model~\cite{simonyan2014very} and the Fast-RCNN~\cite{girshick2015fast} object detection framework. AC-CNN takes an image and its object proposals provided by Selective Search~\cite{uijlings2013selective} as inputs. In recent successful object detection frameworks, such as Fast R-CNN~\cite{girshick2015fast} and Faster R-CNN~\cite{ren2015faster}, features of generated proposals are pooled from the last convolutional layer. Similarly, in our model, an input image first passes the convolutional and pooling layers to generate a feature cube. Then, two context-aware sub-networks (our main contributions) with attention to local and global context are introduced. For convenience of illustration, we term these two sub-networks as multi-scale contextualized sub-network and attention-based contextualized sub-network, respectively. In the following subsections, we explain these two sub-networks  in more details.

\vspace{-2mm}
\subsection{Multi-scale Contextualized Sub-network}
A local view into the neighborhood of a specific proposal can bring some useful cues for inferring the corresponding content. Beyond the original bounding box of a specific proposal, two more scales used to exploit inside and outside contextual information are employed to enhance the feature representation.

We use $\bf{I}$ and $\bf{p}$ to denote the input image and one specific proposal throughout the paper.
The proposal $\bf{p}$ is  encoded by its size $(w, h)$ and coordinates of its center $(c_x, c_y)$, \emph{i.e.}, ${\bf{p}}=(c_x, c_y, w, h)$. To exploit inside and outside contextual information of $\bf{p}$, we crop $\bf{p}$ from the feature cube with three scaling factors: ${\lambda_1}=0.8$, ${\lambda_2}=1.2$ and ${\lambda_3}=1.8$. We denote the features of $\bf{p}$ pooled from  crops of the feature cube with different sizes as $\{{\bf{v}}_{\bf{p}}^i|i=1, 2, 3\}$. Considering implementation, since the output feature should have compatible dimension with that of the pre-trained VGG-16 model on ImageNet, we need to resize the features  to  $7\times7\times512$ before feeding them into the first fully connect layer.

To meet this requirement on the size of input features, we first concatenate  features of crops $\bf{p}$ with different sizes into ${\bf{F}}={concat_{\{i=1,2,3\}}}\{{\bf{v}}_{{\bf{p}}}^i\}$, where $concat$ indicates the operation of concatenating each pooled feature along the channel axis. Each pooled feature is L2 normalized and scaled to match the original amplitudes. Then, we use a $1\times1$ convolution operator to reduce the shape of $\bf{F}$ from $7\times7\times(3\times512)$ to $7\times7\times512$.


Finally, the final feature passes two fully-connected layers to generate the feature representation of $\bf{p}$, which is denoted as ${\bf{F}}_L$. Here the subscript $L$ denotes ``local'' context.

\vspace{-2mm}
\subsection{Attention-based Contextualized Sub-network}
A global view from the entire image can provide useful contextual information for object detection. 
To exploit global context beneficial for object detection and filter out noisy contextual information, we propose to adapt an attention based recurrent model~\cite{sharma2015action} to adaptively identify positive contextual information from the global view. Since input images usually have different sizes, shapes of feature cubes from the last convolutional layer are also different. To calculate the global context with consistent parameters, the feature cube is pooled into a fixed shape of $K \times K \times D$ ($20 \times 20\times 512$ in our experiments). Based on the feature cube, the recurrent model learns an attention map with the shape of $K^2$ to highlight the region that may benefit the object detection from the global view.

We now illustrate how to build the attention model to learn an attention map for extracting global context information. Denote feature slices in the feature cube as $X=[{\bf{x}}_i, \cdots, {\bf{x}}_{K^2}]$, where ${\bf{x}}_i (i=1, \cdots, K^2)$ is with $D$ dimension. The recurrent model is composed of three layers of Long Short Memory (LSTM) units. We adopt the LSTM implementation discussed in the work by Sharma \etal~\cite{sharma2015action}:

\begin{equation}
	\left( \begin{array}{l}
		{{\bf{i}}_t}\\
		{{\bf{f}}_t}\\
		{{\bf{o}}_t}\\
		{{\bf{g}}_t}
	\end{array} \right) = \left( \begin{array}{l}
	\sigma \\
	\sigma \\
	\sigma \\
	{\rm{tanh}}
\end{array} \right)M\left( \begin{array}{l}
{{\bf{h}}_{t - 1}}\\
{{\bf{x}}_t}
\end{array} \right),
	\vspace{-5mm}
\end{equation}

\[\]
\begin{equation}
	{{\bf{c}}_t} = {{\bf{f}}_t} \odot {{\bf{c}}_{t - 1}} + {{\bf{i}}_t} \odot {{\bf{g}}_t},
	\vspace{-3mm}
\end{equation}

\begin{equation}
	{{\bf{h}}_t} = {{\bf{o}}_t} \odot {\bf{tanh}}({{\bf{c}}_t}),
\end{equation}
where ${\bf{i}}_t$, ${\bf{f}}_t$, ${\bf{c}}_t$, ${\bf{o}}_t$ and ${\bf{h}}_t$ denote the input gate, forget gate, cell state, output gate and hidden state of the LSTM, respectively. ${\bf{x}}_t$ is the input to the LSTM at time-step $t$. $M \in {R^{a \times b}}$ is an affine transformation consisting of parameters with $a=d+D$ and $b=4d$, where $d$ is the dimensionality of all of ${\bf{i}}_t$, ${\bf{f}}_t$, ${\bf{c}}_t$, ${\bf{o}}_t$, ${\bf{g}}_t$ and ${\bf{h}}_t$. Besides, $\sigma$ and $\odot$ correspond to the logistic sigmoid activation and the element-wise multiplication, respectively.

At each time-step $t$, the attention model predicts a weighted map ${\bf{l}}_{t+1}$, a softmax over $K \times K$ locations. This is probabilistic estimation about whether the corresponding region in the input image is beneficial for the object classification from the global view. The location softmax at time-step $t$ is computed as follows:


\begin{equation}
	{l}_{t,i}=p\left ( {L}_{t} = i \mid {\bf{h}}_{t-1} \right )= \frac{\exp \left ( W_{i}^{\top }{\bf{h}}_{t-1} \right )}{\sum_{j= 1}^{K\times K}\exp \left ( W_{j}^{\top }{\bf{h}}_{t-1}\right )},\; \; \; \; i\in \{1\cdot \cdot \cdot K^{^{2}}\}
\end{equation}
where ${W}_{i}$ is the weights mapping to the $i^{th}$ element of the location softmax and ${L}_{t}$ is a random variable which takes 1-of-$K^{2}$ value. With these probabilities, the attended feature can be calculated by taking the expectation over the feature slices at different regions following the soft attention mechanism~\cite{bahdanau2014neural}. The feature taken as the input to the LSTM at the next time-step is defined as ${\bf{x}}_t$, which can be calculated as
\begin{equation}
	{\bf{x}}_{t}=\sum_{i=1}^{K^{2}}l_{t,i}{X}_{t,i},
\end{equation}
where ${X}_{t}$ is the feature cube and ${X}_{t,i}$ is the $i^{th}$ slice of the feature cube.

The cell state ${\bf{c}}_t$ and the hidden state ${\bf{h}}_t$ of the LSTM are initialized following the strategy proposed in~\cite{xu2015show} for faster convergence: 

\begin{equation}
	{\bf{c}}_{0}=f_{init,c}\left ( \frac{1}{K^{2}}\sum_{i=1}^{K^{2}}X_{t,i} \right )
\end{equation}

\begin{equation}
	{\bf{h}}_{0}=f_{init,h}\left ( \frac{1}{K^{2}}\sum_{i=1}^{K^{2}}X_{t,i} \right )
\end{equation}
where $f_{init,c}$ and $f_{init,h}$ are two multilayer perceptions. These values are used to calculate the first location softmax ${\bf{l}}_{1}$ which determines the initial input ${\bf{x}}_1$.

\begin{figure}[t]
	\centering
	\includegraphics[scale=0.7]{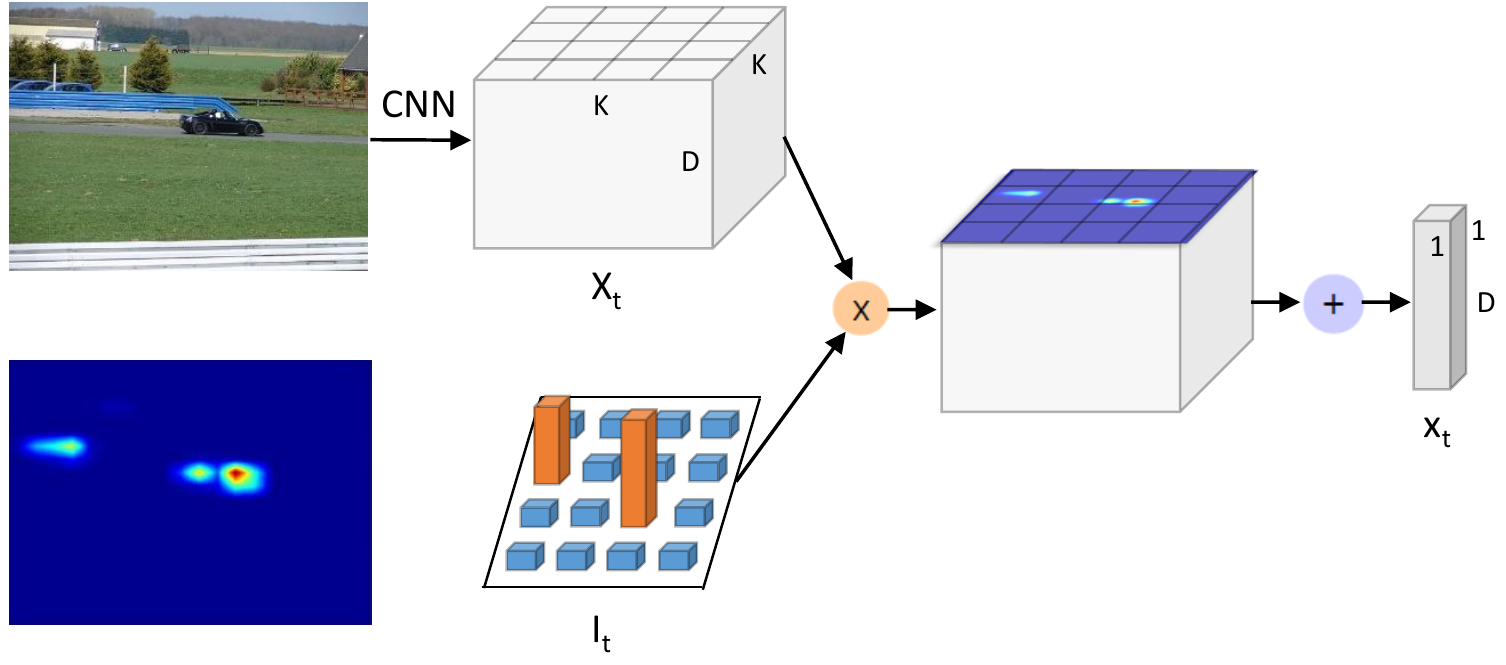}
	\caption{Illustration of how to produce the global attention-based feature. With the attention-based contextualized sub-network, a $K \times K$ attentive location map is calculated to selectively combine the features of all locations into a $1 \times 1 \times D$ global attention-based feature.}
	\label{fig:attention}
	\vspace{-4mm}
\end{figure}

Figure~\ref{fig:attention} illustrates the process of producing the global attention-based feature. It can be observed that a $D$-dimensional global attention-based feature can be obtained by combining features of all locations according to the attentive location map.
Finally, the feature passes through two fully-connected layers to produce the feature representation of $\bf{p}$ with global contextual information, which can be denoted as ${\bf{F}}_G$.

\subsection{Learning with Multi-task Loss Function}
Denote $g \in \{0, 1, \cdots, K\}$ as the ground-truth label out of totally $K+1$ categories (0 indicates $background$). With the obtained feature representations of ${\bf{F}}_L$ and ${\bf{F}}_G$, the loss function for jointly optimizing object classification and bounding box regression could be defined as

\begin{equation}
	J = {J_{cls}}([{\bf{F}}_L,{\bf{F}}_G]) + [g \ge 1]{J_{reg}}({\bf{F}}_L)
\end{equation}
where $[{\bf{F}}_L,{\bf{F}}_G]$ indicates concatenating two features along the channel axis, and $[g \ge 1]$ is equal to 1 when $g \ge 1$ and 0 otherwise. $J_{cls}$ is the cross-entropy loss~\cite{krizhevsky2012imagenet} for the ground-truth class $g$, and $J_{reg}$ is a smooth $L_1$ loss proposed in~\cite{girshick2015fast}. It should be noted that only the feature from the local context net is used for bounding box regression. The corresponding justified experiment is provided in the next section.

\section{Experimental Results}
\subsection{Experimental Setting}
\textbf{Datasets and Evaluation Metrics} We evaluate the proposed AC-CNN on two mainstream datasets: PASCAL VOC 2007 and VOC 2012~\cite{everingham2010the}. These two datasets contain 9,963 and 22,531 images respectively, and they are divided into \emph{train}, \emph{val} and \emph{test} subsets. The model for VOC 2007 is trained based on the \emph{trainval} splits from VOC 2007 (5,011) and VOC 2012 (11,540). The model for VOC 2012 is trained based on all images from VOC 2007 (9,963) and \emph{trainval} split from VOC 2012 (11,540). The evaluation metrics are $Average Precision$ (AP) and $mean ~of AP$ (mAP) complying with the PASCAL challenge protocols.

\noindent\textbf{Implementation Details} The implementation of the proposed AC-CNN adopts VGG-16~\cite{simonyan2014very} as the bottom CNN architecture. Our experiments use the publicly available Fast-RCNN~\cite{girshick2015fast} built on the Caffe~\cite{jia2014caffe} platform. The VGG-16 model is pre-trained on ILSVRC 2012~\cite{deng2009imagenet} classification and localization dataset, which can be downloaded from the Caffe Model Zoo. We fine-tune the proposed AC-CNN based on the pre-trained VGG-16. During the fine-tuning, we use 2 images per mini-batch, which contains 128 selected object proposals. Following Fast R-CNN, 25\% of object proposals are selected as foreground that have Intersection over Union (IoU) overlap with a ground-truth bounding box of larger than 0.5 and the rest are background in each mini-batch. During training, images are horizontally flipped with a probability of 0.5 to augment the training data. All the newly added layers (all the fully-connected layers in the global context net, $1\times1$ convolutional layer in the local context net, the last fully-connected layer for object classification and bounding box regression) are randomly initialized with zero-mean Gaussian distributions with standard deviations of 0.01 and 0.001, respectively.

We run Stochastic Gradient Descent (SGD) for 120K and 150K iterations in total to train the network parameters for VOC 2007 an VOC 2012, respectively. The AC-CNN is trained based on a NVIDIA GeForce Titan X GPU and Intel Core i7-4930K CPU @ 3.40 GHz. The initial learning rate of all layers is set as 0.001 and decreased to one tenth of the current rate of each layer after 50K and 60K iterations for VOC 2007 and VOC 2012, receptively.

\begin{table}[]
		\vspace{-5mm}
	\centering
	\caption{Comparison of object detection results on VOC 2007 \emph{test} and VOC 2012 \emph{test}. }
	\label{tab:voc07}
	\begin{tabular}{lcc|cc}
		\toprule
		\multirow{2}{*}{AP(\%)} & \multicolumn{2}{c}{VOC 2007} & \multicolumn{2}{c}{VOC 2012} \\
		\cmidrule{2-5} 
		& FRCN & AC-CNN  & FRCN & AC-CNN\\
		\midrule
		aeroplane    & 77.0 & $\bf{79.3}$          & 82.3  & $\bf{83.2}$\\
		bicycle      & 78.1 & $\bf{79.4}$          & 78.4  & $\bf{80.8}$\\
		bird         & 69.3 & $\bf{72.5}$          & $\bf{70.8}$  & $\bf{70.8}$\\
		boat         & 59.4 & $\bf{61.0}$          & 52.3  & $\bf{54.9}$\\
		bottle       & 38.3 & $\bf{43.5}$          & 38.7  & $\bf{42.1}$\\
		bus          & $\bf{81.6}$ & 80.1          & 77.8  & $\bf{79.1}$\\
		car          & 78.6 & $\bf{81.5}$          & 71.6  & $\bf{73.4}$\\
		cat          & 86.7 & $\bf{87.0}$         & 89.3  & $\bf{89.7}$\\
		chair        & 42.8 & $\bf{48.5}$          & 44.2  & $\bf{47.0}$\\
		cow          & 78.8 & $\bf{81.9}$          & 73.0  & $\bf{75.9}$\\
		dining table & 68.9 & $\bf{70.7}$          & 55.0  & $\bf{61.8}$\\
		dog          & $\bf{84.7}$ & 83.5          & 87.5  & $\bf{87.8}$\\
		horse        & 82.0 & $\bf{85.6}$          & 80.5  & $\bf{80.9}$\\
		motorbike    & 76.6 & $\bf{78.4}$          & 80.8  & $\bf{81.8}$\\
		person       & 69.9 & $\bf{71.6}$          & 72.0  & $\bf{74.4}$\\
		potted plant & 31.8 & $\bf{34.9}$          & 35.1  & $\bf{37.8}$\\
		sheep        & 70.1 & $\bf{72.0}$          & 68.3  & $\bf{71.6}$\\
		sofa         & $\bf{74.8}$ & 71.4          & 65.7  & $\bf{67.7}$\\
		train        & 80.4 & $\bf{84.3}$          & 80.4  & $\bf{83.1}$\\
		tv/monitor   & 70.4 & $\bf{73.5}$          & 64.2  & $\bf{67.4}$\\
		\midrule
		mAP          & 70.0 & $\bf{72.0}$          & 68.4  & $\bf{70.6}$\\
		\bottomrule
	\end{tabular}
	\vspace{-1mm}
\end{table}

\subsection{Performance Comparisons}
We evaluate the proposed AC-CNN on VOC 2007 and the more challenging VOC 2012 by submitting the results to the publicly evaluation server. Table~\ref{tab:voc07} provides the comparisons of the proposed AC-CNN and Fast-RCNN (FRCN). All the experimental results are based on the model trained on the VOC 2007 \emph{trainval} set merged with the VOC 2012 \emph{trainval} set. 

It can be observed that our method obtains the mAP scores of 72.0\% and 70.6\% on VOC 2007 and VOC 2012, which outperforms the baselines by 2.0\% and 2.2\%, respectively. Our method reaches better detection results on most categories compared with FRCN. Specifically, small or occluded objects often appear in some specific categories such as \emph{bottle}, \emph{chair} and \emph{potted plant}. The improvements on these three challenging categories are 5.2\%, 5.7\% and 3.1\%, which can further validate the effectiveness of the AC-CNN for detecting challenging objects.

\vspace{-2mm}
\subsection{Ablation Analysis on Training Pipeline}
\vspace{-2mm}
To validate the effectiveness of experimental settings in this work, we evaluate some components of AC-CNN.

\vspace{-2mm}
\subsubsection{ Global Context Attention Method}
In the proposed AC-CNN framework, we employ an attention based recurrent model to capture positive contextual information from the global view. To validate the powerful mechanism of LSTMs for global context attending, we compare it with another pooling method, \ie average pooling. We compare these two methods on VOC 2007 \emph{test}, shown in Table~\ref{tab:gp}. It can be observed that the proposed recurrent model performs better than the average pooling scheme. It should be noted that the performance will drop by 0.4\% by adding global context via average pooling compared with ``AC-CNN". The reason may be that not all information from the entire image is useful. By averaging features of all regions, some potential noise may also be introduced, decreasing the detection accuracy. With the recurrent model, an attentive location map can be optimized to highlight those regions that are with the positive effect upon the object detection. Therefore, we use the LSTMs to compute the global context in this work.

\begin{table}
	\centering
	\label{tab:gp}
	\caption{Effect of different global context methods.}
	\begin{tabular}{l|c}
		\toprule
		Global Context    & mAP (\%) \\
		\midrule
		Average Pooling   & 71.6 \\
		Attention-based Pooling  & 72.0 \\
		\bottomrule
	\end{tabular}
	\vspace{-8mm}
\end{table}

\subsubsection{Contributions of Each Sub-network}
The proposed AC-CNN consists of two sub-networks, \ie attention-based contextualized sub-network and multi-scale contextualized sub-network. These two sub-networks are utilized to capture global and local contextual information, respectively. We compare the detection results of different sub-networks on VOC 2007 \emph{test}, shown in Table~\ref{tab:sub}. ``AC-CNN minus L" represents the model of only using attention-based global context information for object detection. ``AC-CNN minus G" represents the model of only using multi-scale local context information for object detection. It can be observed that the performance will drop by 0.6\% if we remove either sub-subnetwork, which can further validate the effectiveness of each component in the proposed AC-CNN.

\begin{table}
	\centering
	\label{tab:sub}
	\caption{Effect of different sub-networks.}
	\begin{tabular}{l|c}
		\toprule
		Model    & mAP (\%) \\
		\midrule
		AC-CNN minus L   & 71.4 \\
		AC-CNN minus G  & 71.4 \\
		AC-CNN          & 72.0 \\
		\bottomrule 
	\end{tabular}
	\vspace{-8mm}
\end{table}
\subsubsection{Effectiveness of Multi-scale Setting}
We compare the detection performance with different scale settings based on the multi-scale contextualized sub-network. It can be observed that the performance will be decreased if we remove any additional scale (0.8 or 1.8). The reason is that inside (or outside) surrounding context of a specific proposal can not be well exploited by removing the corresponding scale, \ie 0.8 (or 1.8). To further validate the effectiveness of contextual information, we also add one more scale (\ie 2.7) for comparison. It can be observed the multi-scale contextualized sub-network will make a further improvement with more scales. However, to balance the detection accuracy and the computational consumption of time and GPU memory, we choose the multi-scale as ``0.8+1.2+1.8" in this work.

\begin{table}
	\centering
	\label{tab:scale}
	\caption{Comparison of multi-scale contextualized sub-network with different scale settings.}
	\begin{tabular}{l|c}
		\toprule
		Model    & mAP (\%) \\
		\midrule
		AC-CNN minus G (0.8 + 1.2)   & 71.3 \\
		AC-CNN minus G (1.2 + 1.8) & 71.1 \\
		AC-CNN minus G (0.8 + 1.2 + 1.8)         & 71.4 \\
		AC-CNN minus G (0.8 + 1.2 + 1.8 + 2.7)         & 71.6 \\
		\bottomrule 
	\end{tabular}
	\vspace{-8mm}
\end{table}

\begin{table}
	\centering
	\label{tab:bbr}
	\caption{Comparison of bounding box regression based on different feature representations.}
	\begin{tabular}{l|c}
		\toprule
		Feature Representation    & mAP (\%) \\
		\midrule
		Local+Global Context  & 71.9 \\
		Local Context   & 72.0 \\
		\bottomrule
	\end{tabular}
	\vspace{-4mm}
\end{table}

\subsubsection{No Global Context for Bounding Box Regression}
We do not use the concatenated feature from both local context net and global context net for bounding box regression. The reason is that bounding box regression is a totally different task compared with object classification. It is excepted that the feature of one specific proposal could represent the difference between its current position and the ground-truth position.  However, by adding the feature from the global view, this difference will be slacked. This may decrease the accuracy of bounding box regression. We compare the results of different settings on VOC 2007 \emph{test}, shown in Table~\ref{tab:bbr}. It can be observed that 0.1\% improvement will be made by removing the global context information for the bounding box regression task.

\begin{figure}[t]
	\centering
	\includegraphics[scale=0.43]{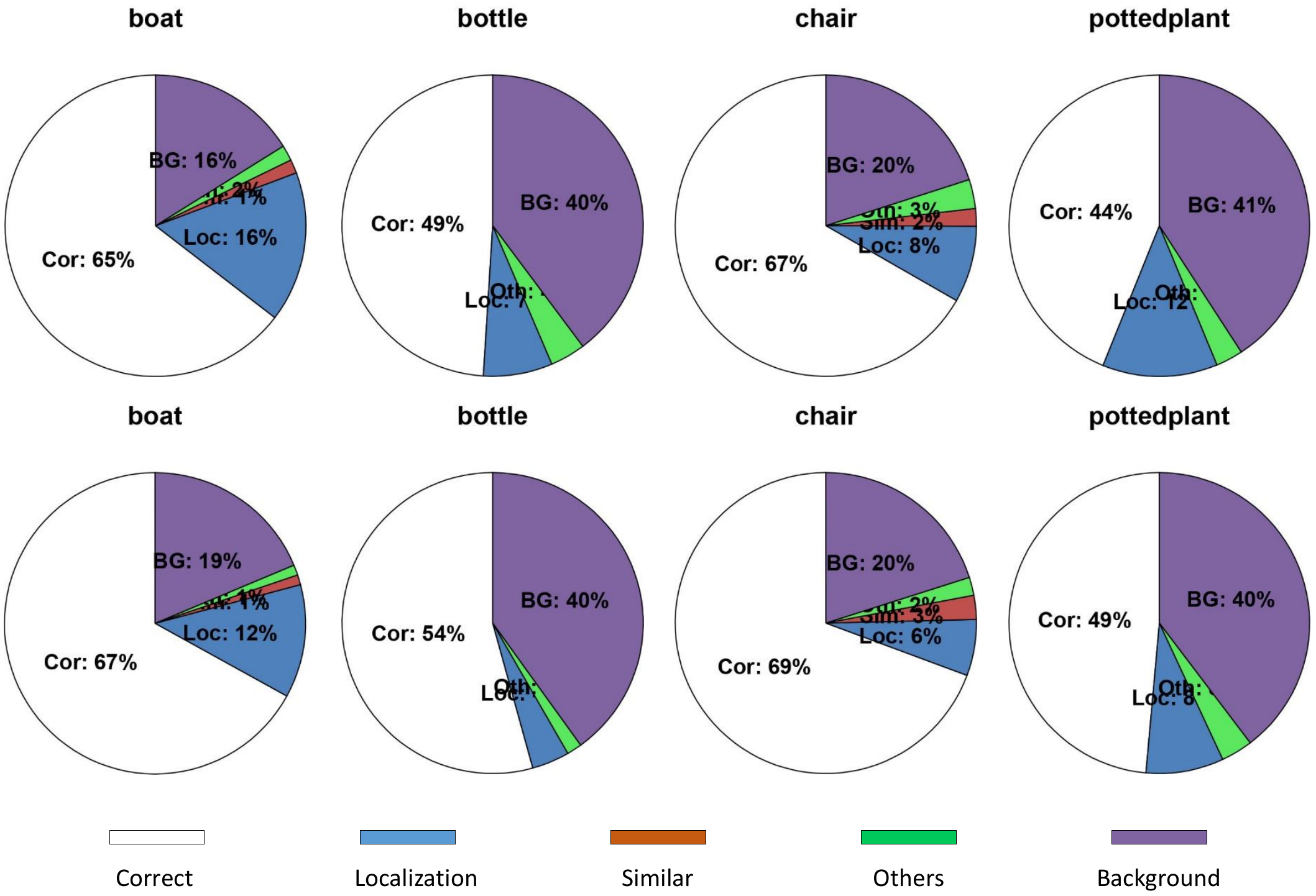}
	\caption{Analysis of top ranked false positives on VOC 2007 \emph{test}. Fraction of top $N$ detections ($N$ is the number of objects in the category) that are correct (Cor), or false positives due to poor localization (Loc), confusion with similar objects (Sim), confusion with other VOC objects (Oth), or confusion with background or unlabeled objects (BG).  We only show the graphs for challenging classes, \ie \emph{boat}, \emph{bottle}, \emph{chair} and \emph{pottedplant}, due to space limitations. \textbf{Top row:} the results of the FRCN model. \textbf{Bottom row:} the results of the proposed AC-CNN model.}
	\label{fig:ana1}
	\vspace{-5mm}
\end{figure} 

\begin{figure}
	\centering
	\includegraphics[scale=0.43]{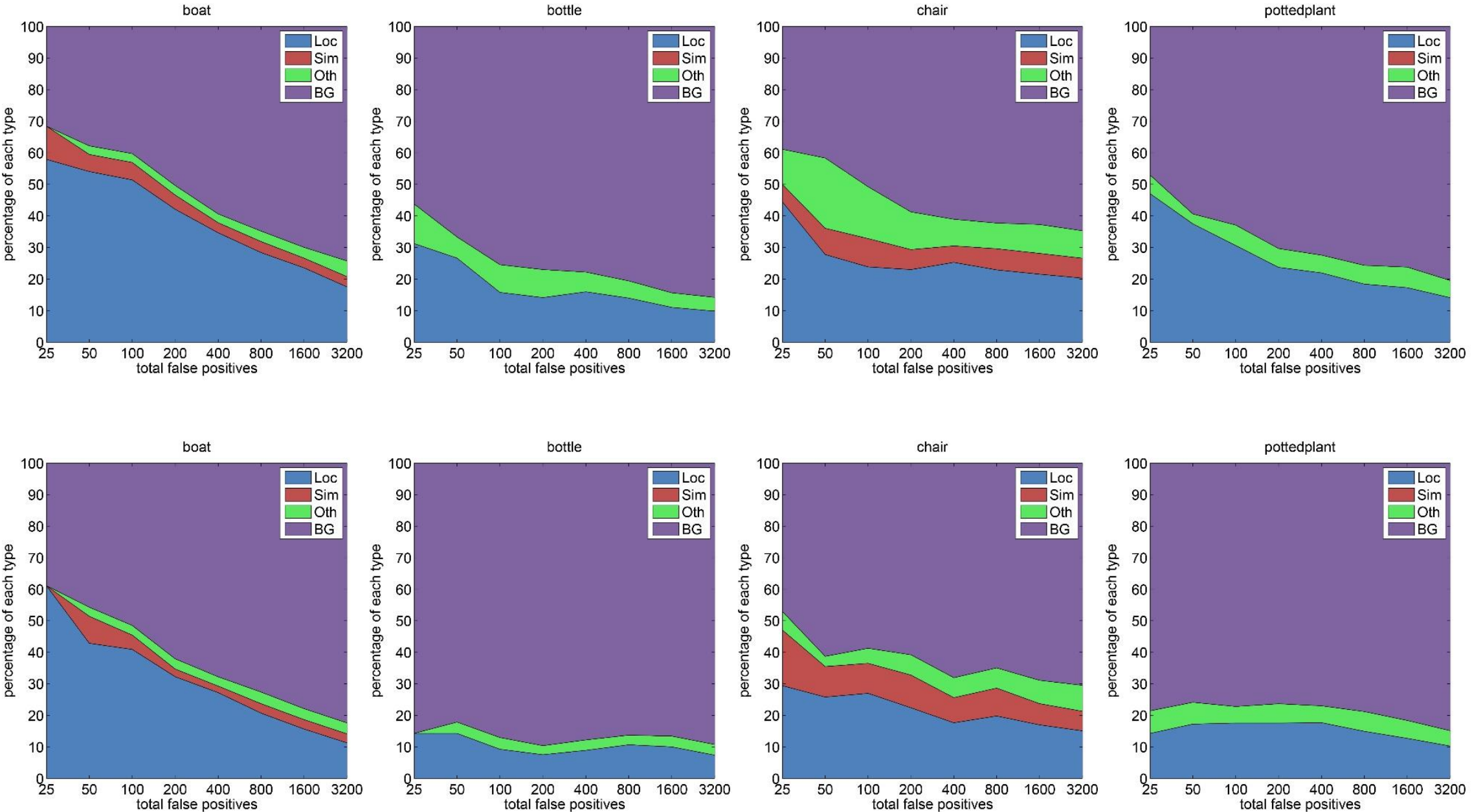}
	\caption{Top ranked false positive types on VOC 2007 \emph{test}. We only show the graphs for challenging classes, \ie \emph{boat}, \emph{bottle}, \emph{chair} and \emph{pottedplant}, due to space limitations. \textbf{Top row:} the results of the FRCN model. \textbf{Bottom row:} the results of the proposed AC-CNN model.}
	\label{fig:ana2}
	\vspace{-2mm}
\end{figure} 

\begin{figure}[t]
	\centering
	\includegraphics[scale=0.44]{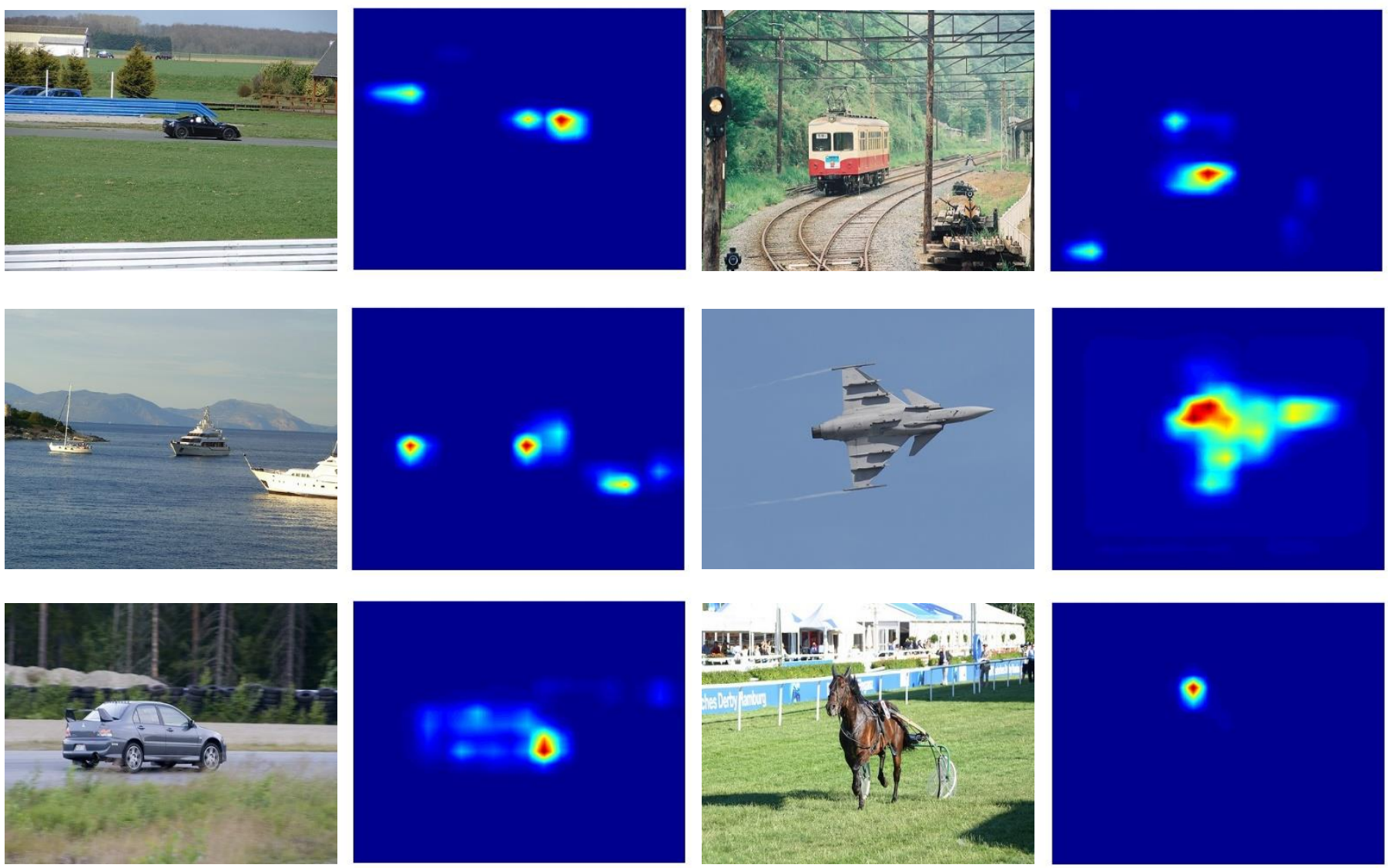}
	\caption{Examples of input images and the corresponding attentive location maps generated by the attention-based contextualized sub-network.}
	\label{fig:gc}
	\vspace{-8mm}
\end{figure} 
\begin{figure}
	\centering
	\includegraphics[scale=0.4]{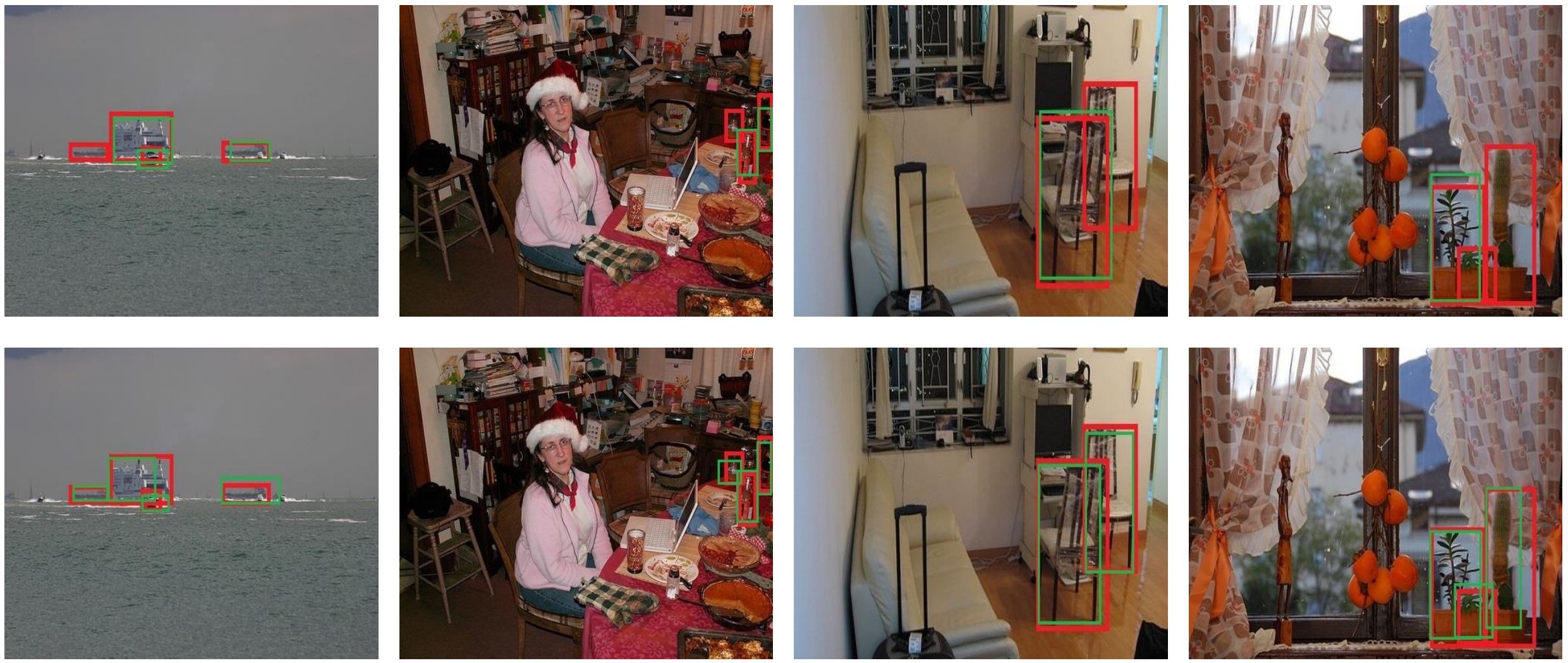}
	\caption{Examples of detection results produced by FRCN and AC-CNN. \textbf{Red} color and \textbf{green} color indicate the ground-truth bounding boxes and the predicted results, respectively.}
	\label{fig:exam}
	\vspace{-4mm}
\end{figure} 

\vspace{-6mm}
\subsubsection{Detection Error Analysis}
The tool of Hoiem \etal~\cite{hoiem2012diagnosing} is employed to analyze the detection errors in this work. As show in Figure~\ref{fig:ana1}, we plot pie charts with the percentage of detections that are correct and false positives due to bad localization, confusion with similar categories and other categories, and confusion with background or unlabeled objects. It can be observed that the proposed AC-CNN model can make a considerable reduction in the percentage of false positives for challenging categories. Specifically, it is well known that small objects usually appear in \emph{bottle} and \emph{pottedplant}. The improvements on these two challenging categories are both 5\%, which can validate the effectiveness of the proposed context-aware method for those objects with small scales. A similar observation can also be deducted from Figure~\ref{fig:ana2} where we plot the top ranked false positive types of the results from FRCN and the proposed AC-CNN.

\vspace{-10mm}
\subsection{Visual Comparison}
\vspace{-2mm}

Figure~\ref{fig:gc} shows some input images and the corresponding attentive maps computed by the attention-based contextualized sub-network. It can be observed that those regions which may benefit the object classification task are highlighted. Some selected detection results on VOC 2007 \emph{test} set are shown in Figure~\ref{fig:exam}. For some small objects, \eg \emph{boat}, \emph{bottle} and \emph{potted plant}, the proposed AC-CNN achieves  better detection results. Besides, for the third group,  the occluded \emph{chair} can also be detected with the AC-CNN model. Therefore, our AC-CNN can effectively advance the Fast-RCNN by incorporating valuable contextual information.

\section{Conclusion}
In this paper, we propose a novel Attention to Context Convolution Neural Networks (AC-CNN) for object detection. Specifically, AC-CNN advances the traditional Fast-CNN to contextualized object detectors, which effectively incorporates the attention-based contextualized sub-network and multi-scale contextualized sub-network into a unified framework. To capture global context, the attention-based contextualized sub-network recurrently generates the attentive location map for the input image by employing the stacked LSTM layers. With the attentive location map, the global contextual features can be produced by selectively combining the feature cubes of all locations. To capture local context, the multi-scale contextualized sub-network exploits the inside and outside contextual information of each specific proposal by scaling the corresponding bounding box with three pre-defined ratios. Extensive experiments on VOC 2007 and VOC 2012 demonstrate that the proposed AC-CNN can make a significant improvement by exploiting contextual information.

\bibliographystyle{splncs}
\bibliography{egbib}
\end{document}